# A Modular and Flexible Architecture for an Integrated Corpus Query System

Oliver Christ

Universität Stuttgart
Institut für maschinelle Sprachverarbeitung (IMS-CL)
Azenbergstr. 12
D 70174 Stuttgart
oli@ims.uni-stuttgart.de

**Abstract**

This paper describes the architecture of an integrated and extensible corpus query system developed at the University of Stuttgart and gives examples of some of the modules realized within this architecture. The modules form the core of a corpus workbench.

Within the proposed architecture, information required for the evaluation of queries may be derived from different knowledge sources (the corpus text, databases, on-line thesauri) and by different means: either through direct lookup in a database or by calling external tools which may infer the necessary information at the time of query evaluation. The information available and the method of information access can be stated declaratively and individually for each corpus, leading to a flexible, extensible and modular corpus workbench.

## 1 Introduction

With the availability of tagged and annotated text corpora, corpora cannot be regarded any more as mere sequences of words. Additionally, more and more linguistic knowledge bases become available and provide additional knowledge about words (MRDs, on-line thesauri like WORDNET [Miller *et al.*, 1993], morphological knowledge bases like the CELEX database [Baayen *et al.*, 1993] , . . . ). When using and querying corpora, all this knowledge should be usable within a corpus query system in order to enable the lexicographer or linguist to express the linguistic properties of the examined phenomenon as precisely as possible (in order to reduce the amount of data which has to be browsed manually), no matter how the knowledge necessary to evaluate the query is stored or by which means it is derived.

When a corpus is thus regarded as a structured object composed of several different knowledge sources, a problem arises because different knowledge sources require possibly different access methods. Furthermore, for many types of information, it is useful not to store the information physically at all but to compute it at the time of query evaluation. For example, bigram tables for large corpora might grow too big to be held online. Automatically assigned part-of-speech tags, on the other hand, might either be stored in a database when they are regarded as "stable" or might be computed at the time of query evaluation by a tagging tool.

Additionally, a corpus query system need not necessarily be used only by human users: a parser might consult a corpus annotated with parse trees (treebank) to disambiguate between several syntactic structures by looking up similar, but disambiguated syntactic patterns; a generator might use a semantically annotated corpus to filter lexical preferences.

These different knowledge sources, access strategies and usage situations are best supported by a hierarchical, modularized system architecture where the single modules can be combined in different ways to adapt the system to various usage situations.

We therefore designed and implemented the following architecture: To abstract as much as possible from the different storage properties, the data access was split between a "logical data access layer", which is independent of data access methods and storage properties, and a "physical data access layer", which is the data-oriented interface to the knowledge sources and which is responsible for data access and network-based corpus data interchange. The adaptation of the system to different usage situations is achieved through different interfaces to the logical access layer, but tools may also request data from the physical layer directly. A general-purpose query language, which treats the whole corpus as a structured knowledge source and allows to express queries involving all knowledge sources declared for a specific corpus (no matter how the knowledge is accessed physically), was added to the logical access layer. This architecture is sketched in figure 1.

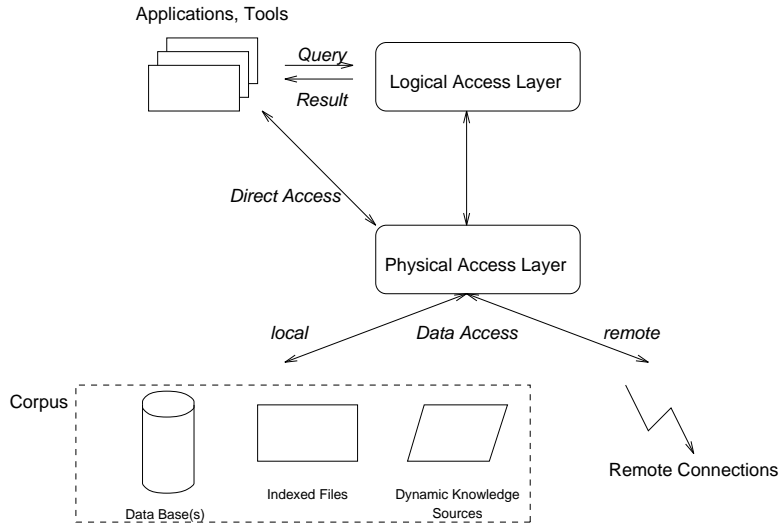

Figure 1: The modular architecture of a flexible corpus query system

In the following sections, these modules are described in more detail. Section 2 outlines the physical layer. In section 3, the logical layer and the query language are described. One usage situation is the interactive use of the query system. For this purpose, presentation and interaction tools have been built which are explained in section 4. In section 5, some directions of our further work are described. The paper ends with a short conclusion in section 6.

## 2 The physical layer

The task of the physical layer is to provide a uniform interface between the logical layer and the files, databases or tools which "store" the information the corpus is built of. The physical layer therefore encapsulates knowledge about file and tool access and provides an interface which is independent of the storage device and the information type (static vs. dynamic). Due to its proximity to the physical corpus representation, the physical layer also provides methods for corpus management, bigram table creation and management, corpus preparation and indexing, frequency counting etc.

Currently, the physical layer supports the following types of corpus annotations:

- *positional attributes* are attributes where a (string) value is assigned to (almost) every corpus position. The sequence of words the corpus text is built of is one example of a positional attribute. Other examples are part-of-speech tags and base forms (see figure 2). An arbitrary number of positional attributes can be assigned to a corpus;

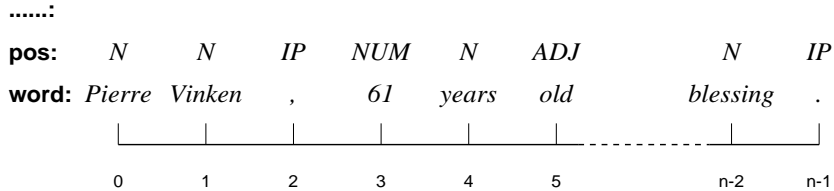

Figure 2: Positional attributes: Values are associated with corpus positions

- *structural attributes* are attributes which capture information about sentence boundaries, article boundaries etc. Currently, recursive structures (like NPs with embedded NPs) cannot be represented. The number of structural attributes is not limited;

- *bigram tables* are related to one of the positional attributes of a corpus and hold information about the absolute number of adjacent occurrences of two values of the attribute within a given window size[1]. Note that, for example, both word bigrams as well as part-of-speech-tag bigrams can be represented;

- *alignment information* can be added to a pair of parallel corpora (which are, roughly speaking, translations of each other) to represent information about corresponding (aligned) ranges (sentences, for example). As in the case of structural attributes, we cannot represent recursive alignments or alignments on more than one level (for example, information about aligned words additionally to aligned sentences);

---

[1] We use the term *attribute value* to denote one element of the list of distinct strings which occur as the values of a positional attribute. In the case of the corpus text, this is the list of distinct words which occur in the corpus.

- finally, *dynamic attributes* are attributes the values of which are not stored physically, but which are computed at query evaluation time by calling external tools, similar to a function call. An arbitrary number of arguments can be declared for a dynamic attribute. When the value of a dynamic attribute is requested, the argument list is filled and an external tool is called. The external tool, then, returns the computed value, which is either a string or an integer value. Neiter indices nor bigram tables can be built for dynamic attributes.

A corpus has to be prepared in a special way before its data can be used by the query system. This preparation step involves character set normalization, tokenization, sentence boundary detection (if required), and – in case of annotated corpora – the partitioning of the different positional attributes (for example, corpus text and part-of-speech tags) into several files. Then, a special one-word-per-line format is produced which is used as input for the construction of the internal corpus representation and the indices[2]. The corpus text itself is not needed any more after transforming it into the internal representation. Details of the internal corpus representation and the encoding steps are described in [Christ, 1994].

After a corpus is encoded, it must be *registered*. This is achieved through a *registry file* which declares the attributes and their types assigned to a corpus. All corpus accessing tools access a corpus only via a symbolic name, which is the file name of the registry file. The tools (and the users) therefore need not know where a corpus is stored in the file system in order to access the data. All relevant information is captured in the registry file.

```
NAME "Hansard corpus (english part)"
ID      hansard-e
HOME /corpora/encoded/hansard-e

ATTRIBUTE word
ATTRIBUTE pos

DYNAMIC ishuman(STRING):INT "/corpora/utils/cmd/wn-hypen '$1' human"

ALIGNED hansard-f            # the french part
```

Figure 3: A small sample registry file

A sample registry file may look as illustrated in figure 3. It declares a corpus `hansard-e` and the directory in which the data can be found. Two positional attributes are assigned to this corpus, `word` and `pos`. Additionally, the dynamic attribute `ishuman` is declared, which takes a string as an argument and returns an integer value (where "0" means "no" and "1" means "yes"). Upon query evaluation, a shell command is executed which consults WORDNET to evaluate whether the argument string may denote a "human object". The corpus is aligned to another corpus, `hansard-f`.

A corpus can be extended after registration. Positional attributes (as well as all other types of attributes) can be added to an existing corpus without need for reindexing existing data.

---

[2]The internal corpus representation we use is inspired by an – unfortunately – unpublished draft paper by Ken W. Church, "A Set of Unix Tools for Processing Large Text Corpora".

For testing purposes, we have implemented a TCP/IP protocol for network-based exchange of corpus data within the physical layer. Through this protocol, it is possible to declare that a given attribute of a corpus (or the whole corpus) is stored on a remote computer. Upon access to remotely stored data, a network connection is built up, access authorization is verified and, if access is granted, the requested data is returned. Through this exchange protocol, it is possible to split corpus data between several computers in the internet. This is useful, for example, to share corpus data between several computers or to run query tools on computers which have too little memory or hard disk space to hold large corpora (although data access is slowed down a lot by remote connections). The remote status of an attribute is hidden within the physical layer, that is, clients of the physical layer do not need to handle remote corpora differently from local data access.

One of the most important "clients" of the physical layer is the logical layer, which is described in the following section. Other clients are tools which do not need to access a corpus through a query language (for example, word list generators or tools which statistically evaluate frequency or bigram counts).

## 3 The logical layer and the query language

The logical layer uses the information provided by the physical layer to parse and evaluate corpus queries given in the query language described below[3]. Within this layer, the set of positional attributes defined on a corpus can be seen as a sequence of entities referred to by corpus positions. These entities may have several attributes, for example the attribute WORD for the "character string" found at a given corpus position, POS for the part-of-speech tag assigned to that word, ROOT for the base form of that word, etc. The query language allows to find sequences of entities where a number of conditions over such attribute-value pairs hold.

Conditions are boolean expressions which involve attribute-value tests, where all positional attributes defined on a corpus can be used. Such a condition may look as follows:

(1)    [word="chair.*" & pos != "N.*"]

When this condition is evaluated against a given corpus position, it is tested whether the value of the word attribute at that corpus position matches (=) the regular expression "chair.*" and the value of the pos attribute does not match (!=) the regular expression "N.*"[4].

A query consists of a regular expression over such conditions. In addition to concatenation of conditions, the other standard regular expression operators are available, like "*" for an arbitrary number of repetitions of the preceding regular expression, "+" for at least one repetition, "?" for optionality, and "|" for disjunction. Parentheses can be used for grouping of expressions. [] is a "wildcard" which matches every corpus position. Additionally, the interval operator $\{n, m\}$ is supported, which denotes at least $n$, but at most $m$ repetitions of the preceding regular expression[5]. Thus, regular expressions are used on the level of attribute values as well as on the level of conditions. Example (1) is already a query, since it is a one-element regular expression.

---

[3] Currently, the logical layer only supports positional, structural and dynamic attributes; access to bigram and alignment attributes has yet to be implemented.

[4] We use the POSIX EGREP syntax for regular expressions. In this standard, the dot "." matches every character and the star "*" matches any (possibly empty) sequence of the last character or regular (sub-)expression. A common error is to write "N*" when all strings beginning with a capital N should be matched, but the regular expression "N*" denotes all strings which entirely consist of a sequence of capital Ns.

When a query is evaluated, the query interpreter computes all matches of the regular expression in the corpus. A match of a query is a "substring" of the corpus, that is, a corpus interval the boundaries of which are the beginning and ending corpus positions of the match. Since regular expressions are used which, in general, may contain repetition operators, these intervals can differ in length. The result of a whole query is the set of matches, that is, a set of corpus intervals.

The following examples illustrate some aspects of the query language. Query (2)

(2)     [pos="JJ.*"] [pos="N.*"] "and|or" [pos="N.*"] [pos="IN" & word != "that"]

returns all corpus intervals which are (adjacent) sequences of an adjective (JJ, JJR, JJS)[6], a noun (NN, NNS), a conjunction, another noun and finally a preposition or subordinating conjunction (IN) which must not be that (in the corpus, that was often tagged as IN, which should be excluded in this query)[7]. When in a condition only the word attribute is accessed (together with the equality operator), the brackets can be omitted. So "and|or" is just an abbreviation for the complete condition [word="and|or"][8].

Dynamic attributes can be accessed in a simple way:

(3)     "kill.*" []? [pos="N.*" & ishuman(word)]

As defined in the sample registry file in figure 3, the dynamic attribute ishuman requires a string argument and returns an integer value which internally is interpreted as "Yes" if the value is 1, and interpreted as "No" if the value is 0. In query 3, ishuman is called with the value of the word attribute of the noun. When the query is evaluated, all matches are computed which are a sequence of a word beginning with kill, followed by an optional, unspecified word (for example, by), and finally followed by a noun for which the consultation of WORDNET gives reason to assume that it may denote a human.

A predefined dynamic attribute is "f", which returns the absolute frequency of its argument in the corpus. To search the "most common human beings who are loved", the following query could be formulated:

(4)     "love.*" []? [pos="N.*" & f(word)>10 & ishuman(word)];

Structural attributes, like sentence boundaries, can be accessed by SGML-like tags:

(5)     [pos="N.*"] [] <s> "She"

This query returns all corpus intervals where a noun, followed by an arbitrary item (which is to match the full stop or other sentence delimiter) occurs in front of a sentence boundary, followed by the word "She".

---

[5]When $m$ is omitted in such an interval, exactly $n$ repetitions are matched.

[6]The corpus on which query (2) was run is a part of the Penn Treebank, which has been tagged with the Penn Treebank POS tagset. See [Marcus *et al.*, 1993] for an explanation of the tags.

[7]Query 2 serves to filter concordances which illustrate the problems of adjective scope and PP-attachment within conjoint noun phrases. For a few matching lines, see figure 4.

[8]The condition "and|or" could as well be expressed as ([word="and"]|[word="or"]), or, abbreviated, as ("and"|"or"). Whereas the latter two expressions use disjunction on the level of conditions (and have to be grouped by parentheses), the expression used in query (2) uses disjunction on the level of attribute values.

Structural attributes like sentence or article boundaries can also be used to limit the search space when repetitions are used. For example, the query

(6)  `"president" []* "said"`

would search the two strings `"president"` and `"said"` separated by an arbitrary number of non-specified items. In general, only those matches which entirely lie within one sentence will be of interest. This can be achieved by using the `within` construct:

(7)  `"president" []* "said" within s;`

Now, the whole match has to lie within the boundaries of one sentence[9]. All structural attributes defined on a corpus can be used as boundary markers (like `<s>`) or in the `within` construct. For example, when the structural attribute `article` was defined on a newspaper corpus, `within article` can be used in queries as well.

An additional, powerful construct of the query language are *label references*, which can be used instead of an attribute value. A condition can be labelled by preceding it with a label name and a colon ("`a:`"), as in (8):

(8)  `a:[pos="N.*"] ...`

Then, in a subsequent condition in the same query, an agreement of attribute values can be expressed:

(9)  `a:[pos="N.*"] []* [pos="PRP" & num=a.num] within s;`

Here, the value of the `num` attribute of the personal pronoun (`PRP`) must be the same as the value of the `num` attribute at the position the label `a` refers to, that is, the value of the number attribute of the noun. The whole match must lie within one sentence. Another example which illustrates the power of label references is the following query:

(10) `a:[pos="N.*"] ([]* [word=a.word] ){2} within s;`

This query returns all intervals where the same noun occurs more than tree times within the same sentence.

The query language implements some additional constructs, which cannot described in detail here. For a full description of the query language, its power and a comparison with other corpus query languages, see [Schulze, 1994].

Query results can be saved in files and reloaded and reviewed in later sessions. The logical layer supports subsequent queries on the result of an earlier query, which can greatly reduce the search space and therefore improves efficiency. For example, in a newspaper corpus, one can first extract all articles of a corpus where a certain syntactic construction is used. Afterwards, this "subcorpus" of articles can be analysed by subsequent queries running only on a part of the original corpus. Additionally, set operators are supported, that is, query results can not only be produced by queries, but also by combining results of earlier queries with union, intersection and difference operators. Through this mechanism, it is possible, for example, to intersect the set of sentences generated by a first query with the set of sentences of a second query to get all sentences where the conditions expressed in both queries hold. Although the same result could possibly be produced by a single query as well, set operators are more user-friendly. In our eyes, searching on the results produced by earlier queries and the possibility to combine query results to new "subcorpora" supports a successive refinement of queries with the gain of efficiency, and allows a stepwise approach to the solution of complex problems.

---

[9]The number of sentences which may "surround" the matched interval can be expessed with a number following the `within` keyword. "`within 2 s`" therefore allows a two-sentence distance.

The result of a query can be postprocessed by different tools for presentation, frequency counting, additional filters etc. The following section describes two simple presentation tools.

## 4 Presentation modules

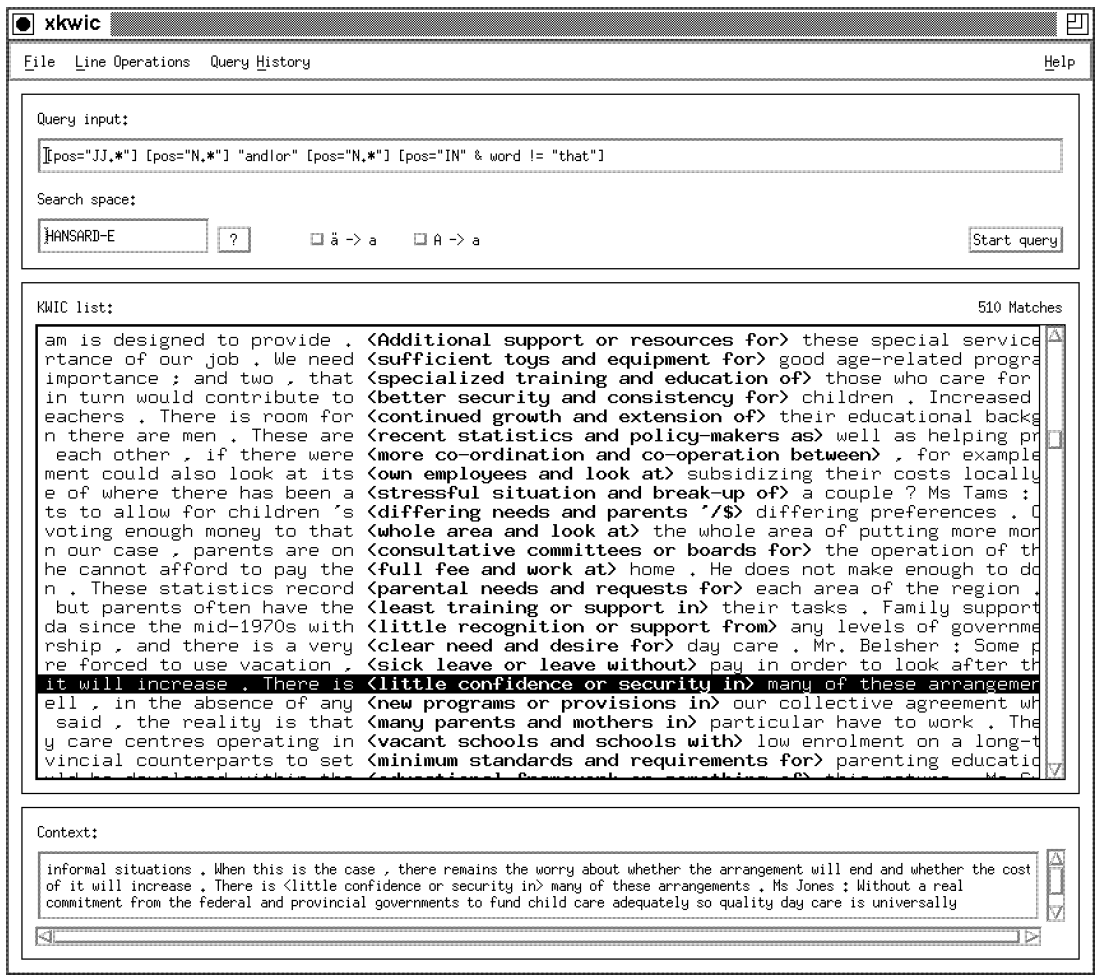

Figure 4: The XKWIC presentation module

A presentation module has the task to display the information returned by a query, suitably formatted for a human user. One instance of such a module is a program called XKWIC which is an X Window System based graphical user interface for displaying key

>     hansard-e: The only way to make and to encourage the responsiveness of child care services to parental and consumer and children 's needs is to encourage competition among those services , to encourage a diversity of services , as indeed exists , so they can reflect the variety of <parental needs and requirements in> this area .
>     hansard-f: Pour que les garderies tiennent véritablement compte des besoins des parents , des consommateurs et des enfants , il faut favoriser la concurrence entre les services , encourager la diversité , telle qu' elle existe , afin de satisfaire aux exigences et aux besoins nombreux des parents dans ce domaine .

Figure 5: Part of the output of a presentation module which uses alignment information

word in context (KWIC) concordances. XKWIC also provides an input area for typing in queries to the logical layer, thus being a general and comfortable interface for corpus work.

Figure 4 shows XKWIC after processing the query displayed in the topmost window within the English part of the HANSARD corpus (the same query as shown in example (2) above). The inverted KWIC line is displayed with a larger context in the bottommost window. XKWIC provides functions to adjust the size of the displayed match context, to sort the query result, to delete single or multiple KWIC lines and a function to write (selected) KWIC lines textually to a file. Additionally, XKWIC supports a simple query history: all queries which are entered are kept in a list which can be saved to a file and reloaded in later sessions. An earlier query can then simply be rerun by clicking on the entry in the query history list. XKWIC is described in more detail in [Christ, 1993].

If corpora are aligned (like the HANSARD corpora) and the alignment was defined in the registry file, another presentation module may be used to display both the query result in the source corpus as well as the aligned portion of the target corpus. The same query result displayed in figure 4 then appears as shown in figure 5 (the matching part of the source corpus is surrounded by angle brackets)[10].

## 5 Further steps

Discussions with users of the query system have shown that it is highly desirable to be able to use parsed corpora in queries. So, one direction of our future work is to design a physical representation of parse trees which allows efficient access and processing and to augment the query language with a construct to refer to this information.

Currently, XKWIC does not support all operations provided by the logical layer, especially the operations on query results (set operations, saving and loading of query results, ...). Therefore, one of our next goals is to integrate the full functionality of the logical layer into a comfortable user interface.

## 6 Conclusions

The modular architecture of the corpus query system described in this paper has several advantages:

---

[10] It would be possible to integrate the second module into XKWIC, but this hasn't yet been done.

- several knowledge sources can be added to individual corpora. The knowledge they provide can then be used in corpus queries;

- knowledge sources or annotations can be added to a corpus without the necessity of reindexing existing data;

- through a flexible data model, the information necessary to evaluate the query may be derived from different sources, can be computed at query evaluation time or can be gathered from remote computers;

- through the separation of storage, evaluation and presentation tasks into different modules, the whole system can be adapted to different usage situations.

The flexibility achieved by this architecture, together with the power of the query language, provide the linguist or lexicographer with an extensible and comfortable corpus workbench which allows the querying of corpora with much more precision than within frameworks based only on the corpus text. This leads to more specific queries and results, reducing the amount of data which has to be browsed manually.